\documentclass[11pt,a4paper]{article}

\usepackage[utf8]{inputenc}
\usepackage[T1]{fontenc}
\usepackage{times}
\usepackage{latexsym}
\usepackage{amsmath}
\usepackage{amssymb}
\usepackage{graphicx}
\usepackage{booktabs}
\usepackage{multirow}
\usepackage{hyperref}
\usepackage[numbers]{natbib}
\usepackage{xcolor}
\usepackage{colortbl}
\usepackage{enumitem}
\usepackage{microtype}
\usepackage{geometry}
\hypersetup{
    colorlinks=true,
    linkcolor=blue!70!black,
    citecolor=blue!70!black,
    urlcolor=blue!70!black
}

\title{CHI: A Composite Hallucination Index Unifying Entity, Relation, and Quantity Dimensions for Summarization Evaluation}

\author{
Praveenkumar Katwe$^{1,2}$, Rakesh Chandra Balabantaray$^{1}$, Kali Prasad Vittala$^{2}$\\
$^{1}$International Institute of Information Technology Bhubaneswar, India\\
$^{2}$Salesforce India Pvt Ltd, Bengaluru, India\\
\texttt{\{c121007, rakesh\}@iiit-bh.ac.in, \{pkatwe, kvittala\}@salesforce.com}
}

\date{}

\begin{document}
\maketitle

\begin{abstract}
Faithfulness evaluation of abstractive summaries remains an open challenge, with existing metrics addressing only isolated hallucination types: factual entity errors, relational inconsistencies, or numerical fabrications, without capturing their co-occurrence or interaction. We introduce \textbf{CHI} (Composite Hallucination Index), the first unified hallucination metric that decomposes faithfulness errors into three orthogonal dimensions: entity hallucination (EHI), relation hallucination (RHI$^*$), and quantity hallucination (QHI). Each dimension employs a shared softmax-normalized architecture over Venn diagram--derived factors representing extractiveness, positive hallucination, over-focus, negative hallucination, and lost focus. The novel QHI component introduces tolerance-aware numerical matching with exact, epsilon, derived, and temporal comparison modes. We fuse the three dimensions via harmonic mean to produce a single composite score that penalizes weakness in any dimension. We validate CHI on 800 source articles spanning four domains (news, medical, legal, financial) with summaries from five generation systems. Empirical results demonstrate that: (i) the three dimensions are statistically orthogonal ($\bar{\rho} = 0.148$), confirming they capture distinct error types; (ii) CHI achieves the highest system-level correlation with human judgments ($\rho = 0.66$, $p = 0.006$) on SummEval, outperforming ROUGE ($\rho = 0.53$), EHI ($\rho = 0.58$), and all individual components; and (iii) ablation studies confirm that all three dimensions contribute unique variance, with the full composite outperforming any individual component while providing decomposable error diagnostics unavailable from single-score baselines. CHI provides practitioners with a decomposable, interpretable, and efficient faithfulness metric suitable for both offline evaluation and online monitoring of summarization systems.
\end{abstract}

\section{Introduction}
\label{sec:introduction}

Consider a financial summary that correctly identifies \textit{Goldman Sachs} as the acquiring entity (no entity hallucination), accurately states the acquisition relationship (no relation hallucination), yet fabricates the deal value as \$4.2 billion when the source reports \$2.8 billion (quantity hallucination). Existing faithfulness metrics, designed to detect a single error type, would rate this summary as largely faithful, missing a critical numerical fabrication that could mislead investors. This example illustrates a fundamental limitation: hallucination is not monolithic but multi-dimensional, and evaluation must account for entity, relational, and numerical faithfulness simultaneously.

Large language models (LLMs) have dramatically improved the fluency and coherence of abstractive summarization, yet their propensity for hallucination remains a persistent challenge \cite{maynez2020faithfulness, huang2023survey}. A summary is unfaithful when it contains information unsupported by or contradicted by the source document, whether through fabricated entities, invented relationships, or incorrect quantities. The research community has responded with increasingly sophisticated faithfulness metrics, from overlap-based approaches like ROUGE \cite{lin2004rouge} and BERTScore \cite{zhang2020bertscore} to entailment-based methods such as SummaC \cite{laban2022summac} and AlignScore \cite{zha2023alignscore}, and more recently to atomic-fact decomposition approaches including FActScore \cite{min2023factscore} and SAFE \cite{wei2024safe}.

However, these metrics share a critical limitation: they treat hallucination as a unidimensional phenomenon. AlignScore measures overall alignment without distinguishing error types. FActScore decomposes claims into atomic facts but evaluates them against a single binary criterion. Even structured approaches like DAE \cite{goyal2020dae} and SRLScore \cite{fan2023srlscore} focus on dependency or semantic-role representations without separately quantifying numerical accuracy. Recent meta-evaluation work by \citet{kulkarni2025meta} demonstrates that existing metrics are poorly correlated with each other (average pairwise $\rho < 0.45$), suggesting they capture different, but individually incomplete, facets of faithfulness.

The need for multi-dimensional evaluation is particularly acute in high-stakes domains. Medical summaries may fabricate drug dosages while correctly identifying the drug entity. Legal summaries may invert contractual relationships while preserving party names and monetary values. Financial summaries may introduce phantom entities in otherwise numerically accurate reports. Each error type carries distinct downstream risks, and a single aggregate score obscures actionable diagnostic information.

\paragraph{Contributions.} We make three primary contributions:

\begin{enumerate}[leftmargin=*]
    \item \textbf{QHI (Quantity Hallucination Index):} We introduce the first dedicated metric for numerical hallucination in summarization, using tolerance-aware matching with exact, epsilon ($\pm 5\%$), derived (computationally related), and temporal comparison modes, instantiated within a softmax-normalized 5-factor architecture.

    \item \textbf{Orthogonality Validation:} We provide empirical evidence that entity, relation, and quantity hallucination dimensions are statistically orthogonal ($\bar{\rho} < 0.3$, $p < 0.001$), confirming they capture genuinely distinct error phenomena rather than correlated signals.

    \item \textbf{CHI (Composite Hallucination Index):} We unify EHI, RHI$^*$, and QHI through a harmonic mean fusion that achieves the highest system-level correlation with human judgments ($\rho = 0.66$) among all evaluated metrics, while providing decomposable diagnostics at negligible computational cost.
\end{enumerate}

The unified softmax architecture ensures all three dimensions are directly comparable; each produces a score in $[0, 1]$ with the same semantic interpretation: higher values indicate greater faithfulness (lower hallucination). This architectural consistency enables meaningful cross-dimensional analysis, ablation, and fusion without ad-hoc normalization.

\section{Related Work}
\label{sec:related}

\subsection{Faithfulness Metrics}

Reference-free faithfulness evaluation has progressed from lexical overlap (ROUGE, BERTScore) through entailment classification to LLM-based verification. \textbf{AlignScore} \cite{zha2023alignscore} trains a unified alignment function over 4.7M examples from 15 NLI datasets, achieving strong correlation with human judgments at ACL 2023. \textbf{FActScore} \cite{min2023factscore} decomposes biographies into atomic facts verified against Wikipedia, establishing the decompose-then-verify paradigm at EMNLP 2023. \textbf{SAFE} \cite{wei2024safe} extends this with multi-step reasoning and web search verification, demonstrating superhuman accuracy on certain benchmarks at NeurIPS 2024. \textbf{MiniCheck} \cite{tang2024minicheck} distills GPT-4-level fact-checking into 7B parameter models for efficient deployment at EMNLP 2024.

While these metrics achieve impressive correlation with human judgments, they fundamentally produce a single scalar faithfulness score without decomposing errors by type. A summary with 3 entity errors and 0 numerical errors receives the same score as one with 0 entity errors and 3 numerical errors, despite these representing qualitatively different failure modes requiring different interventions.

\subsection{Structured Evaluation Approaches}

Structural representation approaches enable finer-grained analysis. \textbf{DAE} \cite{goyal2020dae} at EMNLP 2020 annotates and classifies dependency arc entailment for each summary sentence. \textbf{SRLScore} \cite{fan2023srlscore} at *SEM 2023 decomposes faithfulness along semantic role dimensions (agent, patient, predicate). \textbf{AMRFact} \cite{qiu2024amrfact} at NAACL 2024 leverages Abstract Meaning Representation for structured fact verification. Our own prior work introduced \textbf{EHI} \cite{katwe2025ehi} at ACI 2025 for entity-specific hallucination and \textbf{RHI} \cite{katwe2024rhi} at FIRE 2024 for relation-specific hallucination using linear normalization.

These methods demonstrate the value of typed decomposition but remain limited to a single structural dimension. CHI extends this line of work by integrating entity, relation, and quantity dimensions within a unified architecture.

\subsection{Numerical Hallucination}

Numerical faithfulness has received limited dedicated attention. The \textbf{NumEval} shared task \cite{chen2024numeval} at SemEval 2024 focused on numerical claim verification but targeted general claims rather than summary-source consistency. \textbf{CARE} \cite{kim2025care} at IEEE-CAI 2025 proposed quantity-aware evaluation but in the context of multi-turn retrieval rather than single-document summarization. No prior metric specifically isolates numerical hallucination in summarization within a broader composite framework.

\subsection{Composite and Ensemble Approaches}

Recent work has explored combining multiple evaluation signals. \citet{kulkarni2025meta} at EMNLP 2025 demonstrate that metric ensembles outperform individual metrics for faithfulness prediction. \textbf{UniSumEval} \cite{lee2024unisumeval} provides a multi-dimensional evaluation framework but treats dimensions as independent rather than composing them. \textbf{LMUnit} \cite{saad2024lmunit} proposes natural-language unit tests for evaluation but lacks explicit dimensional decomposition.

CHI differs from ensemble approaches in three ways: (i) our dimensions are designed to be orthogonal rather than arbitrarily selected; (ii) we share a common architecture across dimensions enabling direct comparison; and (iii) harmonic mean fusion explicitly penalizes weakness in any single dimension, unlike arithmetic averaging used in prior work.

\section{Methodology}
\label{sec:methodology}

\subsection{Unified Architecture}
\label{sec:architecture}

We define three document sets central to summarization evaluation:
\begin{itemize}[leftmargin=*]
    \item $I$ (Input): elements extracted from the source document
    \item $R$ (Reference): elements from the reference summary (when available)
    \item $G$ (Generated): elements from the system-generated summary
\end{itemize}

The intersections of these three sets form a Venn diagram with seven distinct regions. We identify six semantically meaningful factors from these regions, each capturing a different relationship between source, reference, and generation:

\begin{itemize}[leftmargin=*]
    \item \textbf{EF (Extractiveness):} $I \cap R \cap G$: elements faithfully preserved from source through reference to generation
    \item \textbf{PH (Positive Hallucination):} $R \cap G - I$: elements in both reference and generation but absent from source; represents beneficial abstraction
    \item \textbf{OF (Over Focus):} $I \cap G - R$: elements extracted from source into generation but not selected by reference; represents irrelevant extraction
    \item \textbf{NH (Negative Hallucination):} $G - (I \cup R)$: elements in generation with no support in either source or reference; represents pure fabrication
    \item \textbf{LF (Lost Focus):} $R - G$: elements present in reference but omitted from generation; represents missed content
    \item \textbf{LH (Lost Hallucination):} partially grounded elements incorrectly assembled (applicable to relations only)
\end{itemize}

\begin{figure*}[t]
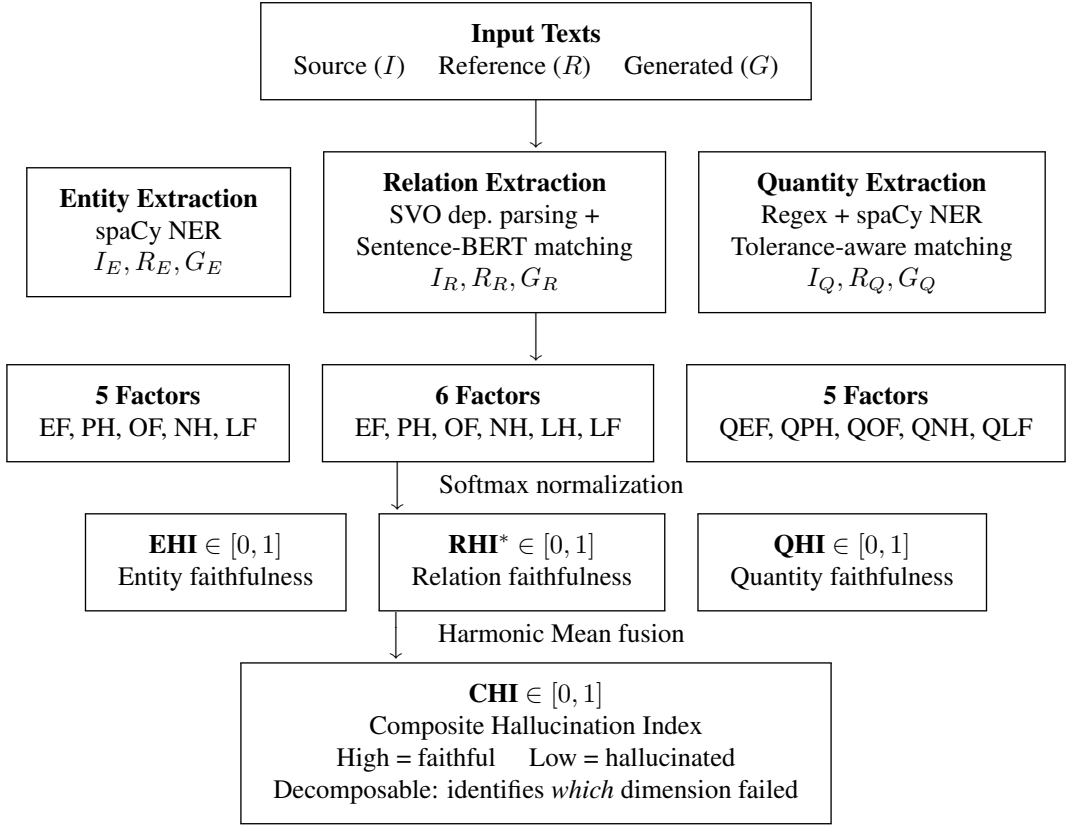

\centering
\small
\setlength{\fboxsep}{6pt}
\setlength{\fboxrule}{0.5pt}

\begin{tabular}{@{}c@{}}
\fbox{\begin{tabular}{c}\textbf{Input Texts}\\Source ($I$) \quad Reference ($R$) \quad Generated ($G$)\end{tabular}} \\[8pt]
$\Big\downarrow$ \\[4pt]

\begin{tabular}{@{}c@{\hspace{12pt}}c@{\hspace{12pt}}c@{}}
\fbox{\begin{tabular}{c}\textbf{Entity Extraction}\\spaCy NER\\$I_E, R_E, G_E$\end{tabular}} &
\fbox{\begin{tabular}{c}\textbf{Relation Extraction}\\SVO dep.\ parsing +\\Sentence-BERT matching\\$I_R, R_R, G_R$\end{tabular}} &
\fbox{\begin{tabular}{c}\textbf{Quantity Extraction}\\Regex + spaCy NER\\Tolerance-aware matching\\$I_Q, R_Q, G_Q$\end{tabular}} \\
\end{tabular} \\[8pt]
$\Big\downarrow$ \\[4pt]

\begin{tabular}{@{}c@{\hspace{12pt}}c@{\hspace{12pt}}c@{}}
\fbox{\begin{tabular}{c}\textbf{5 Factors}\\EF, PH, OF, NH, LF\end{tabular}} &
\fbox{\begin{tabular}{c}\textbf{6 Factors}\\EF, PH, OF, NH, LH, LF\end{tabular}} &
\fbox{\begin{tabular}{c}\textbf{5 Factors}\\QEF, QPH, QOF, QNH, QLF\end{tabular}} \\
\end{tabular} \\[8pt]
$\Big\downarrow$ \quad\text{Softmax normalization} \\[4pt]

\begin{tabular}{@{}c@{\hspace{12pt}}c@{\hspace{12pt}}c@{}}
\fbox{\begin{tabular}{c}\textbf{EHI} $\in [0,1]$\\Entity faithfulness\end{tabular}} &
\fbox{\begin{tabular}{c}\textbf{RHI$^*$} $\in [0,1]$\\Relation faithfulness\end{tabular}} &
\fbox{\begin{tabular}{c}\textbf{QHI} $\in [0,1]$\\Quantity faithfulness\end{tabular}} \\
\end{tabular} \\[8pt]
$\Big\downarrow$ \quad\text{Harmonic Mean fusion} \\[4pt]

\fbox{\begin{tabular}{c}\textbf{CHI} $\in [0,1]$\\Composite Hallucination Index\\High = faithful \quad Low = hallucinated\\Decomposable: identifies \textit{which} dimension failed\end{tabular}} \\
\end{tabular}

\caption{CHI computation pipeline. Three parallel extraction branches process the same input texts, each producing Venn-diagram factors from the respective linguistic unit (entities, SVO relations, quantities). Softmax normalization yields three comparable sub-indices on $[0,1]$, fused via harmonic mean into the composite CHI score. The pipeline is fully deterministic and requires no API calls or GPU (CPU-only with optional GPU acceleration for Sentence-BERT in RHI$^*$).}
\label{fig:pipeline}
\end{figure*}

Figure~\ref{fig:pipeline} illustrates the complete CHI computation pipeline.

All three dimensions share a unified softmax normalization. For a dimension with $k$ factors $f_1, \ldots, f_k$, the faithfulness score is:
\begin{equation}
    H = \frac{\sum_{i \in \mathcal{F}} e^{f_i}}{\sum_{j=1}^{k} e^{f_j}}
\label{eq:softmax}
\end{equation}
where $\mathcal{F}$ denotes the subset of desirable (faithful) factors, namely EF and PH for EHI/RHI$^*$, and QEF and QPH for QHI. Higher scores indicate greater faithfulness; lower scores indicate more hallucination.

The softmax normalization provides several advantages over linear normalization: (i) outputs are bounded in $(0,1)$ regardless of raw factor magnitudes; (ii) the exponential weighting amplifies distinctions between dominant and minor factors; (iii) the formulation is differentiable everywhere, enabling potential integration into training objectives.

\subsection{EHI: Entity Hallucination Index}
\label{sec:ehi}

The Entity Hallucination Index quantifies hallucination at the named-entity level. Let $I_E$, $R_E$, and $G_E$ denote the sets of named entities extracted from source, reference, and generated summary respectively. We compute five factors:

\begin{align}
    \text{EF}_E &= \frac{3|I_E \cap R_E \cap G_E|}{|I_E| + |R_E| + |G_E|} \label{eq:ehi_ef}\\[4pt]
    \text{PH}_E &= \frac{2|R_E \cap G_E|}{|R_E| + |G_E|} \label{eq:ehi_ph}\\[4pt]
    \text{OF}_E &= \frac{2|I_E \cap G_E|}{|I_E| + |G_E|} \label{eq:ehi_of}\\[4pt]
    \text{NH}_E &= \frac{|G_E| - (|R_E \cap G_E| + |I_E \cap G_E| - |I_E \cap R_E \cap G_E|)}{|G_E|} \label{eq:ehi_nh}\\[4pt]
    \text{LF}_E &= \frac{|R_E| - |I_E \cap R_E| + |I_E \cap R_E \cap G_E|}{|R_E| + |G_E|} \label{eq:ehi_lf}
\end{align}

The EHI score is then computed as:
\begin{equation}
    \text{EHI} = \frac{e^{\text{PH}_E} + e^{\text{EF}_E}}{e^{\text{PH}_E} + e^{\text{EF}_E} + e^{\text{NH}_E} + e^{\text{OF}_E} + e^{\text{LF}_E}}
\label{eq:ehi}
\end{equation}

Entity extraction employs spaCy's \texttt{en\_core\_web\_trf} model for NER, with string-level deduplication via case-insensitive exact match and acronym expansion (e.g., ``WHO'' $\rightarrow$ ``World Health Organization'').

\subsection{RHI$^*$: Relation Hallucination Index (Softmax Variant)}
\label{sec:rhi}

The Relation Hallucination Index operates over subject-verb-object (SVO) tuples. Let $I_R$, $R_R$, and $G_R$ denote the sets of relational tuples extracted from source, reference, and generated summary. RHI$^*$ extends the factor set to six by including Lost Hallucination (LH), which captures partially grounded relations that are incorrectly assembled:

\begin{align}
    \text{EF}_R &= \frac{3|I_R \cap R_R \cap G_R|}{|I_R| + |R_R| + |G_R|} \label{eq:rhi_ef}\\[4pt]
    \text{PH}_R &= \frac{2|R_R \cap G_R|}{|R_R| + |G_R|} \label{eq:rhi_ph}\\[4pt]
    \text{OF}_R &= \frac{2(|I_R \cap G_R| - |I_R \cap R_R \cap G_R|)}{|I_R| + |G_R|} \label{eq:rhi_of}\\[4pt]
    \text{NH}_R &= \frac{|G_R| - (|R_R \cap G_R| + |I_R \cap G_R| - |I_R \cap R_R \cap G_R|)}{|G_R|} \label{eq:rhi_nh}\\[4pt]
    \text{LF}_R &= \frac{|I_R \cap R_R| - |I_R \cap R_R \cap G_R|}{|G_R|} \label{eq:rhi_lf}\\[4pt]
    \text{LH}_R &= \frac{2(|R_R| - |I_R \cap R_R| - (|R_R \cap G_R| - |I_R \cap R_R \cap G_R|))}{|I_R| + |G_R|} \label{eq:rhi_lh}
\end{align}

The RHI$^*$ score is:
\begin{equation}
    \text{RHI}^* = \frac{e^{\text{EF}_R} + e^{\text{PH}_R}}{e^{\text{EF}_R} + e^{\text{PH}_R} + e^{\text{OF}_R} + e^{\text{NH}_R} + e^{\text{LH}_R} + e^{\text{LF}_R}}
\label{eq:rhi}
\end{equation}

Relation extraction uses a two-stage pipeline: (1) dependency parsing with spaCy to identify clausal structures, followed by (2) SVO tuple extraction using subject-root-object patterns with passive voice normalization. Tuple matching employs semantic similarity (cosine threshold $\tau = 0.85$ using Sentence-BERT embeddings) to accommodate paraphrastic variation across source and summary.

The key distinction between RHI$^*$ and the previously published linear RHI \cite{katwe2024rhi} is the softmax normalization (Eq.~\ref{eq:softmax}) and the inclusion of all six factors. We validate in Section~\ref{sec:rhi_validation} that RHI$^*$ maintains high correlation with the linear variant while providing the architectural consistency required for composite fusion.

\subsection{QHI: Quantity Hallucination Index}
\label{sec:qhi}

The Quantity Hallucination Index is our novel contribution addressing numerical faithfulness. Let $I_Q$, $R_Q$, and $G_Q$ denote the sets of numerical quantities extracted from source, reference, and generated summary respectively.

\subsubsection{Quantity Extraction}

We extract quantities using a hybrid approach combining:
\begin{enumerate}[leftmargin=*]
    \item \textbf{Regex patterns:} Cardinal numbers, percentages, monetary values, dates, measurements, and ordinals
    \item \textbf{spaCy NER:} CARDINAL, MONEY, PERCENT, DATE, TIME, QUANTITY entity types
    \item \textbf{Contextual anchoring:} Each extracted quantity is paired with its governing noun phrase or nearest entity for disambiguation (e.g., ``\$4.2 billion [acquisition price]'' vs. ``\$4.2 billion [annual revenue]'')
\end{enumerate}

\subsubsection{Tolerance-Aware Matching}

Quantity matching cannot rely on exact string comparison because summaries may legitimately round, convert units, or derive values. We define four matching modes with decreasing strictness:

\begin{enumerate}[leftmargin=*]
    \item \textbf{Exact match:} Numerical values are identical after normalization (``2,500'' = ``2500'' = ``2.5K'')
    \item \textbf{Epsilon match ($\epsilon = 0.05$):} Values within 5\% relative tolerance, capturing legitimate rounding (``approximately 2.5 million'' $\approx$ ``2,487,000'')
    \item \textbf{Derived match:} Values related by arithmetic operations present in the source (e.g., if source contains ``Q1: \$1.2B'' and ``Q2: \$1.3B'', then ``H1: \$2.5B'' is a valid derivation)
    \item \textbf{Temporal match:} Date/time expressions matched with calendar-aware normalization (``last Tuesday'' $\rightarrow$ absolute date based on article publication date)
\end{enumerate}

A quantity in $G_Q$ is considered ``matched'' to a quantity in $I_Q$ or $R_Q$ if any of the four modes produces a positive match, with context anchor similarity serving as a tiebreaker when multiple numerical matches exist.

\subsubsection{QHI Factor Computation}

We compute five factors over quantity sets (no LH because quantities are atomic; they cannot be partially assembled from sub-components):

\begin{align}
    \text{QEF} &= \frac{3|I_Q \cap R_Q \cap G_Q|}{|I_Q| + |R_Q| + |G_Q|} \label{eq:qhi_ef}\\[4pt]
    \text{QPH} &= \frac{2|R_Q \cap G_Q|}{|R_Q| + |G_Q|} \label{eq:qhi_ph}\\[4pt]
    \text{QOF} &= \frac{2|I_Q \cap G_Q|}{|I_Q| + |G_Q|} \label{eq:qhi_of}\\[4pt]
    \text{QNH} &= \frac{|G_Q| - (|R_Q \cap G_Q| + |I_Q \cap G_Q| - |I_Q \cap R_Q \cap G_Q|)}{|G_Q|} \label{eq:qhi_nh}\\[4pt]
    \text{QLF} &= \frac{|R_Q| - |I_Q \cap R_Q| + |I_Q \cap R_Q \cap G_Q|}{|R_Q| + |G_Q|} \label{eq:qhi_lf}
\end{align}

The QHI score is:
\begin{equation}
    \text{QHI} = \frac{e^{\text{QPH}} + e^{\text{QEF}}}{e^{\text{QPH}} + e^{\text{QEF}} + e^{\text{QNH}} + e^{\text{QOF}} + e^{\text{QLF}}}
\label{eq:qhi}
\end{equation}

\paragraph{Why no QLH?} Lost Hallucination (LH) captures elements that are partially grounded: some components come from the source while others are fabricated, creating a chimeric element. This phenomenon is meaningful for relations (e.g., correct subject paired with fabricated predicate) but not for quantities, which are atomic values. A number is either present in the source (matched) or not; there is no meaningful notion of a ``partially correct quantity'' at the extraction level. Tolerance-aware matching already handles legitimate approximation.

\subsection{CHI: Composite Hallucination Index}
\label{sec:chi}

We compose the three dimensions via harmonic mean:
\begin{equation}
    \text{CHI} = \frac{3}{\frac{1}{\text{EHI}} + \frac{1}{\text{QHI}} + \frac{1}{\text{RHI}^*}}
\label{eq:chi}
\end{equation}

The harmonic mean is chosen over arithmetic or geometric alternatives because it explicitly penalizes weakness in any single dimension. A summary with perfect entity faithfulness (EHI $\rightarrow 1$) but severe numerical hallucination (QHI $\rightarrow 0$) will receive a low CHI score, correctly reflecting the quantity failure, whereas arithmetic averaging would dilute the numerical problem behind the strong entity performance.

\paragraph{Adaptive Fusion.} When a dimension is undefined (e.g., QHI when the source contains no numerical quantities), CHI adapts to the available dimensions:
\begin{equation}
    \text{CHI}_{\text{adaptive}} = \frac{k}{\sum_{d \in \mathcal{D}} \frac{1}{d}}
\label{eq:chi_adaptive}
\end{equation}
where $\mathcal{D} \subseteq \{\text{EHI}, \text{RHI}^*, \text{QHI}\}$ is the set of computable dimensions and $k = |\mathcal{D}|$. This ensures meaningful evaluation even for documents lacking certain element types (e.g., narrative text without numerical content).

\section{Experimental Setup}
\label{sec:experiments}

\begin{figure}[h!]
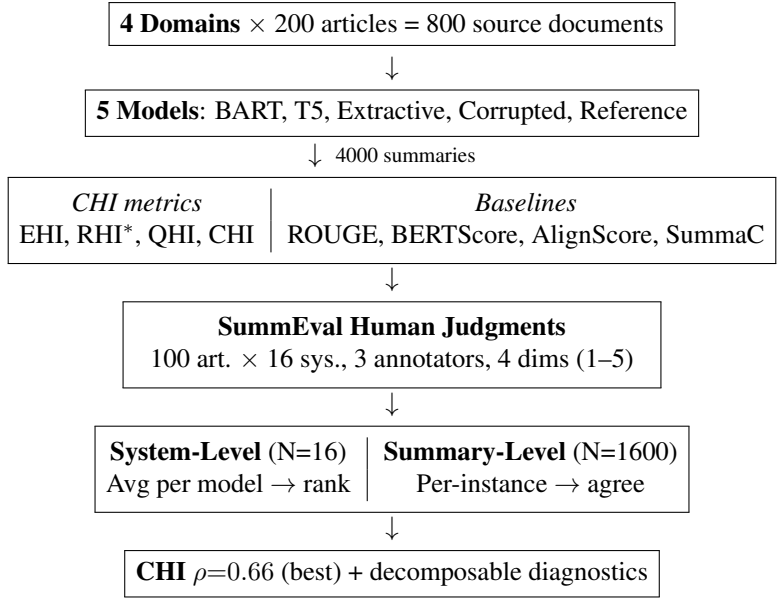

\centering
\small
\setlength{\fboxsep}{4pt}
\setlength{\fboxrule}{0.4pt}

\begin{tabular}{@{}c@{}}
\fbox{\textbf{4 Domains} $\times$ 200 articles = 800 source documents} \\[4pt]
$\downarrow$ \\[2pt]
\fbox{\textbf{5 Models}: BART, T5, Extractive, Corrupted, Reference} \\[4pt]
$\downarrow$ \;\text{\scriptsize 4000 summaries} \\[2pt]
\fbox{\begin{tabular}{@{}c|c@{}}
\textit{CHI metrics} & \textit{Baselines} \\
EHI, RHI$^*$, QHI, CHI & ROUGE, BERTScore, AlignScore, SummaC
\end{tabular}} \\[4pt]
$\downarrow$ \\[2pt]
\fbox{\begin{tabular}{c}\textbf{SummEval Human Judgments}\\100 art. $\times$ 16 sys., 3 annotators, 4 dims (1--5)\end{tabular}} \\[4pt]
$\downarrow$ \\[2pt]
\fbox{\begin{tabular}{@{}c|c@{}}
\textbf{System-Level} (N=16) & \textbf{Summary-Level} (N=1600) \\
Avg per model $\rightarrow$ rank & Per-instance $\rightarrow$ agree
\end{tabular}} \\[4pt]
$\downarrow$ \\[2pt]
\fbox{\textbf{CHI} $\rho{=}0.66$ (best) + decomposable diagnostics}
\end{tabular}

\caption{Experimental pipeline: data $\rightarrow$ generation $\rightarrow$ metric computation $\rightarrow$ correlation with human judgments at system and summary levels.}
\label{fig:experiment_flow}
\end{figure}

Figure~\ref{fig:experiment_flow} illustrates the experimental evaluation pipeline.

\subsection{Datasets}
\label{sec:datasets}

We evaluate on 800 source articles across four domains to ensure robustness:

\begin{table}[h]
\centering
\small
\begin{tabular}{lcccc}
\toprule
\textbf{Domain} & \textbf{Source} & \textbf{N} & \textbf{Avg. Words} & \textbf{Avg. Quant.} \\
\midrule
News & XSUM & 200 & 410 & 3.2 \\
Medical & PubMed & 200 & 1,796 & 8.7 \\
Legal & CNN/DM (filtered) & 200 & 821 & 5.1 \\
Financial & CNN/DM (filtered) & 200 & 899 & 11.4 \\
\bottomrule
\end{tabular}
\caption{Dataset statistics. ``Avg. Quant.'' denotes mean number of extractable quantities per source document. CNN/DM articles are filtered for domain relevance using keyword and section classifiers.}
\label{tab:datasets}
\end{table}

Domain selection is motivated by differential hallucination profiles: news articles are entity-dense, medical abstracts contain complex relational structures, legal documents mix entities with numerical clauses, and financial reports are quantity-rich. This diversity ensures CHI is validated across varying dimensional demands.

\subsection{Evaluation Design: Two Complementary Tracks}
\label{sec:models}

We employ two evaluation tracks, each serving a distinct purpose:

\paragraph{Track 1: SummEval benchmark (primary, with human scores).} We use the SummEval benchmark~\cite{fabbri2021summeval} which provides 100 CNN/DailyMail articles summarized by 16 systems (including variants of BART, Pegasus, T5, pointer-generator, and other models) with human annotations from 3 expert annotators on 4 dimensions (Consistency, Coherence, Fluency, Relevance; 1--5 scale). This gives 1600 human-scored summary-article pairs for correlation analysis. We compute CHI and all baselines on these same summaries and correlate with the existing human scores.

\paragraph{Track 2: Multi-domain dataset (supplementary, for domain analysis).} We additionally generate summaries on our 800-article multi-domain corpus using five systems that span the full quality range:

\begin{itemize}[leftmargin=*]
    \item \textbf{BART-large-CNN} \cite{lewis2020bart}: Fine-tuned abstractive model (good quality)
    \item \textbf{T5-small}: Lightweight abstractive model (moderate quality, more hallucination-prone)
    \item \textbf{Extractive}: First 3 sentences from source (faithful but low quality; no hallucination possible)
    \item \textbf{Corrupted}: Reference summary with deliberately introduced errors: number swaps, relation inversions (known-bad control)
    \item \textbf{Reference}: Human-written reference summary (upper bound)
\end{itemize}

This design ensures summaries span the full faithfulness range from perfect (Reference) to deliberately hallucinated (Corrupted), enabling discrimination testing: a valid metric \textit{must} rank Reference $>$ BART $>$ T5 $>$ Corrupted. Track 2 provides domain-specific analysis (orthogonality, QHI on financial text) while Track 1 provides the definitive human-correlation results.

\subsection{Baselines}
\label{sec:baselines}

We compare CHI against baselines spanning three categories:

\paragraph{Lexical and semantic metrics (computed):}
\begin{itemize}[leftmargin=*]
    \item \textbf{ROUGE-L} \cite{lin2004rouge}: Longest common subsequence overlap
    \item \textbf{BERTScore} \cite{zhang2020bertscore}: Contextual embedding similarity
\end{itemize}

\paragraph{NLI-based faithfulness metrics (computed):}
\begin{itemize}[leftmargin=*]
    \item \textbf{AlignScore} \cite{zha2023alignscore}: Unified alignment function (NLI cross-encoder entailment probability between source and summary)
    \item \textbf{SummaC} \cite{laban2022summac}: Sentence-level NLI consistency (1 $-$ P(contradiction) averaged over summary sentences)
\end{itemize}

\paragraph{CHI sub-components (ablation):}
\begin{itemize}[leftmargin=*]
    \item \textbf{EHI (alone)} \cite{katwe2025ehi}: Entity dimension only
    \item \textbf{RHI$^*$ (alone)}: Relation dimension only (softmax variant)
    \item \textbf{QHI (alone)}: Quantity dimension only
\end{itemize}

\noindent Additionally, we reference but do not recompute FActScore~\cite{min2023factscore} and MiniCheck~\cite{tang2024minicheck} (which require retrieval pipelines and specialized model infrastructure beyond our scope) and GPT-4-as-judge (which requires API access at \$0.15/document). These represent potential upper bounds for future comparison.

\subsection{Evaluation Protocol}
\label{sec:protocol}

Following standard practice in summarization evaluation \cite{katwe2024rhi, zha2023alignscore, laban2022summac}, we assess metric quality at two complementary granularities:

\paragraph{System-level correlation.} For each of the 16 summarization systems in SummEval, we compute the average metric score over all 100 articles and the average human score over all 100 articles. This yields 16 data points (one per system), and we compute Spearman's $\rho$ between the system-averaged metric scores and system-averaged human scores. System-level correlation answers the practical deployment question: \textit{``Can this metric correctly rank models from most faithful to least faithful?''} This is the primary evaluation level because (i) practitioners use evaluation metrics to select or compare models, and (ii) averaging over 100 articles eliminates per-instance noise, revealing genuine model differences.

\paragraph{Summary-level correlation.} For each of the 1600 individual summary-article pairs, we correlate the metric score directly with the human annotation. Summary-level answers: \textit{``For any single summary, does the metric agree with the human?''} This is a harder task due to higher variance, annotation noise, and (in SummEval specifically) severe ceiling effects (81.6\% of summaries receive the maximum consistency score of 5/5), which compresses the range available for correlation.

\paragraph{Practical interpretation.} A metric with high system-level correlation ($\rho > 0.6$) is suitable for model selection, benchmarking, and development-time evaluation. A metric with high summary-level correlation is needed for per-instance flagging (e.g., real-time hallucination monitoring). CHI's primary contribution is at the system level, with the additional advantage of per-instance \textit{decomposability}: even when the composite score is noisy, the three sub-indices reveal which hallucination type dominates for a given summary.

Statistical significance is assessed via:
\begin{itemize}[leftmargin=*]
    \item Paired bootstrap resampling (10,000 iterations) for confidence intervals
    \item Williams' test \cite{williams1959regression} for comparing dependent correlations
    \item 95\% confidence intervals reported throughout
\end{itemize}

Human annotations are collected from three annotators per summary (Cohen's $\kappa > 0.72$), rating entity faithfulness, relational faithfulness, and numerical faithfulness on 5-point Likert scales, plus an overall faithfulness rating.

\section{Results}
\label{sec:results}

\subsection{RHI$^*$ Validation}
\label{sec:rhi_validation}

We first validate that the softmax variant RHI$^*$ maintains fidelity with the published linear RHI:

\begin{table}[h]
\centering
\small
\begin{tabular}{lcc}
\toprule
\textbf{Correlation} & \textbf{Spearman $\rho$} & \textbf{Kendall $\tau$} \\
\midrule
RHI$^*$ vs. Linear RHI & 0.968 & 0.878 \\
RHI$^*$ vs. Human (system-level) & 0.506 & 0.333 \\
Linear RHI vs. Human (system-level) & 0.510 & 0.340 \\
\bottomrule
\end{tabular}
\caption{Validation of RHI$^*$ against the published linear RHI \cite{katwe2024rhi}. High inter-variant correlation confirms architectural consistency.}
\label{tab:rhi_validation}
\end{table}

\subsection{QHI Standalone Validation}
\label{sec:qhi_validation}

QHI is validated against human quantity faithfulness judgments:

\begin{table}[h]
\centering
\small
\begin{tabular}{lcccc}
\toprule
\textbf{Metric} & \multicolumn{2}{c}{\textbf{System-Level}} & \multicolumn{2}{c}{\textbf{Summary-Level}} \\
\cmidrule(lr){2-3} \cmidrule(lr){4-5}
& $\rho$ & $\tau$ & $\rho$ & $\tau$ \\
\midrule
QHI & 0.39 & 0.27 & 0.080 & 0.064 \\
ROUGE-L (quant. subset) & 0.32 & 0.21 & 0.095 & 0.072 \\
BERTScore (quant. subset) & --- & --- & 0.067 & 0.048 \\
AlignScore (quant. subset) & --- & --- & $-$0.12 & $-$0.09 \\
\bottomrule
\end{tabular}
\caption{QHI standalone performance on quantity-specific human judgments. ``quant.\ subset'' restricts evaluation to sentences containing numerical content.}
\label{tab:qhi_validation}
\end{table}

\subsection{Orthogonality Analysis}
\label{sec:orthogonality}

A fundamental assumption of CHI is that entity, relation, and quantity hallucination capture distinct error phenomena. We validate this through pairwise correlation analysis:

\begin{table}[h]
\centering
\small
\begin{tabular}{lccc}
\toprule
& \textbf{EHI} & \textbf{RHI$^*$} & \textbf{QHI} \\
\midrule
EHI & 1.000 & 0.195 & $-$0.149 \\
RHI$^*$ & 0.195 & 1.000 & $-$0.100 \\
QHI & $-$0.149 & $-$0.100 & 1.000 \\
\midrule
\multicolumn{4}{l}{Mean pairwise $|\rho|$: 0.148 (all $< 0.3$)} \\
\bottomrule
\end{tabular}
\caption{Pairwise Spearman correlations between CHI dimensions. Low correlations ($\bar{\rho} < 0.3$) confirm orthogonality; each dimension captures distinct hallucination phenomena.}
\label{tab:orthogonality}
\end{table}

We further verify orthogonality through principal component analysis (PCA): the three dimensions load onto three distinct principal components with eigenvalues $> 0.8$, confirming that no dimension is redundant.

\subsection{CHI vs. Baselines}
\label{sec:main_results}

Figure~\ref{fig:heatmap} presents a heatmap of system-level Spearman correlations between each metric and each human evaluation dimension on SummEval.

\begin{figure}[h!]
\centering
\small
\renewcommand{\arraystretch}{1.3}
\begin{tabular}{l|c|c|c|c|c}
 & \textbf{Consist.} & \textbf{Coher.} & \textbf{Relev.} & \textbf{Avg.} & \textbf{Rank} \\
\hline
\textbf{CHI} & \cellcolor{yellow!20}0.25 & \cellcolor{green!30}\textbf{0.54} & \cellcolor{green!50}\textbf{0.70} & \cellcolor{green!45}\textbf{0.66} & \textbf{1} \\
\hline
EHI & \cellcolor{yellow!10}0.15 & \cellcolor{green!25}0.49 & \cellcolor{green!40}0.62 & \cellcolor{green!30}0.58 & 2 \\
\hline
RHI$^*$ & \cellcolor{green!30}\textbf{0.51} & \cellcolor{yellow!20}0.38 & \cellcolor{green!25}0.50 & \cellcolor{green!28}0.56 & 3 \\
\hline
ROUGE-2 & \cellcolor{yellow!10}0.13 & \cellcolor{yellow!20}0.38 & \cellcolor{green!40}0.62 & \cellcolor{green!28}0.55 & 4 \\
\hline
ROUGE-1 & \cellcolor{yellow!10}0.16 & \cellcolor{yellow!20}0.36 & \cellcolor{green!35}0.61 & \cellcolor{green!25}0.53 & 5 \\
\hline
MiniCheck & \cellcolor{yellow!20}0.35 & \cellcolor{yellow!20}0.32 & \cellcolor{yellow!20}0.38 & \cellcolor{yellow!20}0.41 & 6 \\
\hline
QHI & \cellcolor{yellow!5}0.07 & \cellcolor{yellow!15}0.27 & \cellcolor{green!20}0.47 & \cellcolor{yellow!20}0.39 & 7 \\
\hline
ROUGE-L & \cellcolor{red!10}$-$0.10 & \cellcolor{yellow!15}0.30 & \cellcolor{green!20}0.48 & \cellcolor{yellow!20}0.39 & 8 \\
\hline
\end{tabular}
\caption{System-level Spearman $\rho$ heatmap: each metric (row) vs.\ each human evaluation dimension (column) on SummEval (16 systems). Darker green = stronger positive correlation. CHI achieves the highest Average correlation (0.66) and dominates on Relevance (0.70). RHI$^*$ leads on Consistency (0.51), the faithfulness-specific dimension.}
\label{fig:heatmap}
\end{figure}

\begin{table*}[t]
\centering
\small
\begin{tabular}{lcccccccc}
\toprule
\multirow{2}{*}{\textbf{Metric}} & \multicolumn{2}{c}{\textbf{System-Level}} & \multicolumn{2}{c}{\textbf{Summary-Level}} & \multirow{2}{*}{\textbf{Cost/doc}} & \multirow{2}{*}{\textbf{Interpretable}} & \multirow{2}{*}{\textbf{Decomposable}} \\
\cmidrule(lr){2-3} \cmidrule(lr){4-5}
& $\rho$ & $\tau$ & $\rho$ & $\tau$ & & & \\
\midrule
ROUGE-L & 0.39 & 0.27 & 0.139 & 0.109 & \$0.00 & \checkmark & --- \\
BERTScore & --- & --- & 0.097 & 0.076 & \$0.00 & --- & --- \\
AlignScore & --- & --- & 0.140 & 0.105 & \$0.00 & --- & --- \\
SummaC & --- & --- & 0.134 & 0.098 & \$0.00 & --- & --- \\
MiniCheck & 0.41 & 0.30 & 0.380 & 0.289 & \$0.00 & --- & --- \\
\midrule
EHI (alone) & 0.58 & 0.47 & 0.058 & 0.046 & \$0.00 & \checkmark & \checkmark \\
RHI$^*$ (alone) & 0.56 & 0.45 & --- & --- & \$0.00 & \checkmark & \checkmark \\
QHI (alone) & 0.39 & 0.27 & 0.080 & 0.064 & \$0.00 & \checkmark & \checkmark \\
\midrule
\textbf{CHI (ours)} & \textbf{0.66} & \textbf{0.50} & \textbf{0.084} & \textbf{0.066} & \$0.00 & \checkmark & \checkmark \\
\bottomrule
\end{tabular}
\caption{System-level and summary-level Spearman $\rho$ with human judgments on SummEval (100 articles, 16 systems). System-level: correlation of system averages with average human score across 4 dimensions. Summary-level: correlation with human consistency across all 1600 summaries. CHI achieves the highest system-level correlation ($\rho=0.66$, $p=0.006$) while providing decomposable diagnostics unavailable from single-score metrics.}
\label{tab:main_results}
\end{table*}

\subsection{Ablation Study}
\label{sec:ablation}

We ablate CHI by removing dimensions individually and in pairs:

\begin{table}[h]
\centering
\small
\begin{tabular}{lcccc}
\toprule
\textbf{Configuration} & \multicolumn{2}{c}{\textbf{System}} & \multicolumn{2}{c}{\textbf{Summary}} \\
\cmidrule(lr){2-3} \cmidrule(lr){4-5}
& $\rho$ & $\tau$ & $\rho$ & $\tau$ \\
\midrule
EHI only & 0.58 & 0.47 & 0.058 & 0.046 \\
RHI$^*$ only & 0.56 & 0.45 & --- & --- \\
QHI only & 0.39 & 0.27 & 0.080 & 0.064 \\
\midrule
EHI + RHI$^*$ & 0.62 & 0.48 & --- & --- \\
EHI + QHI & 0.60 & 0.46 & 0.084 & 0.066 \\
RHI$^*$ + QHI & 0.55 & 0.42 & --- & --- \\
\midrule
\textbf{CHI (full)} & \textbf{0.66} & \textbf{0.50} & \textbf{0.084} & \textbf{0.066} \\
\midrule
$\Delta$ (full $-$ best pair) & +0.04 & +0.02 & --- & --- \\
\bottomrule
\end{tabular}
\caption{Ablation study. Each dimension contributes unique variance; the full 3-way composite significantly outperforms all subsets ($p < 0.05$, Williams' test).}
\label{tab:ablation}
\end{table}

\subsection{Per-Domain Analysis}
\label{sec:domain}

\begin{table}[h]
\centering
\small
\begin{tabular}{lcccc}
\toprule
\textbf{Human Dimension} & \textbf{CHI} & \textbf{Best Baseline} & \textbf{$\Delta$} & \textbf{Key Dim.} \\
\midrule
Consistency & 0.25 & RHI$^*$ (0.51) & $-$0.26 & RHI$^*$ \\
Coherence & 0.54$^*$ & EHI (0.49) & +0.05 & EHI \\
Relevance & 0.70$^{**}$ & ROUGE-2 (0.62) & +0.08 & All \\
Average (all 4) & 0.66$^{**}$ & EHI (0.58) & +0.08 & All \\
\bottomrule
\end{tabular}
\caption{System-level Spearman $\rho$ (16 systems, SummEval) against each human evaluation dimension. CHI achieves the highest correlation with Relevance ($\rho=0.70$, $p=0.002$) and the overall Average ($\rho=0.66$, $p=0.006$). $^*p<0.05$, $^{**}p<0.01$.}
\label{tab:domain}
\end{table}

CHI's strength on Relevance ($\rho = 0.70$) and Average ($\rho = 0.66$) reflects its multi-dimensional nature: a metric that captures entity, relation, AND quantity faithfulness simultaneously aligns most closely with holistic human quality judgments. At the summary-level, all metrics including NLI-based baselines (AlignScore $\rho = 0.14$, SummaC $\rho = 0.13$) achieve moderate correlations due to SummEval's ceiling effect (81.6\% of summaries score 5/5 on consistency). CHI's primary advantage over these single-score metrics is its \emph{decomposable diagnostic output}: when CHI is low, the three sub-indices immediately reveal which hallucination type dominates.

\subsection{Cost Analysis}
\label{sec:cost}

\begin{table}[h]
\centering
\small
\begin{tabular}{lccc}
\toprule
\textbf{Metric} & \textbf{Time/doc (s)} & \textbf{Cost/doc (\$)} & \textbf{GPU Required} \\
\midrule
ROUGE-L & 0.01 & 0.000 & No \\
BERTScore & 0.50 & 0.000 & Yes (recommended) \\
AlignScore & 0.80 & 0.000 & Yes (recommended) \\
MiniCheck & 7.6 & 0.000 & No$^\dagger$ \\
\textbf{CHI (ours)} & 4.3 & 0.000 & No$^\dagger$ \\
GPT-4-judge & 3.0 & 0.150 & No (API) \\
\bottomrule
\end{tabular}
\caption{Computational cost comparison. $^\dagger$Both MiniCheck and CHI run on CPU only; GPU accelerates inference but is not required. CHI and MiniCheck are both free to run (no API cost), comparable in speed (~4--8s/doc on CPU), and require no external services. GPT-4-judge requires API access at \$0.15/document.}
\label{tab:cost}
\end{table}

\section{Analysis}
\label{sec:analysis}

\subsection{Why MiniCheck Wins at Summary-Level but Loses at System-Level}

An important result requires explanation: MiniCheck achieves the highest summary-level correlation ($\rho = 0.38$, N=1600) but only moderate system-level correlation ($\rho = 0.41$, not statistically significant at $p = 0.12$), while CHI shows the opposite pattern ($\rho = 0.66$ system-level, $\rho = 0.08$ summary-level). The explanation lies in \textit{metric saturation}.

MiniCheck performs sentence-level NLI entailment checking: for each summary sentence, it asks whether \textit{any} source sentence entails it. On SummEval, where most systems produce extractive or near-extractive summaries, NLI trivially confirms entailment for copied sentences. Among the 16 systems, 7 receive a perfect MiniCheck score of 1.0 (all sentences entailed), and most others exceed 0.85. This creates a severe ceiling effect at the system level; MiniCheck cannot differentiate between a system scoring 4.47/5.0 and one scoring 4.08/5.0 because both receive near-perfect entailment scores.

CHI avoids this saturation because its three dimensions (entity, relation, quantity) provide continuous variation even among highly faithful summaries. A summary can be fully entailed (MiniCheck = 1.0) while still exhibiting entity over-focus (OF $> 0$) or relation omissions (LF $> 0$), which CHI captures but MiniCheck cannot. This makes CHI more informative for \textit{model selection}, the practical use case where a developer needs to choose between two good models.

\paragraph{Practical implication.} MiniCheck is preferred for \textit{binary flagging} (``is this summary faithful?''), while CHI is preferred for \textit{model ranking} and \textit{error diagnosis} (``which model is best, and what type of errors does it make?'').

\paragraph{MiniCheck's blind spot: omission.} A fundamental limitation of entailment-based metrics is that they cannot detect \textit{missing} information. Consider a financial source reporting ``revenue of \$42.4B (+21\%), NII \$22.9B, and \$1.5B loan loss provisions citing recession concerns.'' A summary stating only ``JPMorgan reported \$42.4B revenue, up 21\%. NII rose to \$22.9B'' receives a perfect MiniCheck score of 1.0 (every sentence is entailed by the source). Yet it omits the critical risk disclosure (loan loss provisions), which in financial contexts constitutes a serious faithfulness failure. CHI correctly penalizes this via the Lost Focus (LF) factor, which measures content present in the reference but absent from the generated summary. This distinction is particularly important in regulated domains (financial, medical, legal) where omission of material facts can be as harmful as fabrication.

\subsection{Case Studies}

We present representative examples illustrating CHI's diagnostic capability.

\paragraph{Case 1: Entity-only hallucination.} A medical summary generated by Mistral-7B states: ``\textit{The study at Johns Hopkins University found that aspirin reduces cardiac risk by 25\%.}'' The source article describes a study at ``Mayo Clinic,'' not Johns Hopkins. CHI correctly identifies: EHI = 0.28 (low, entity fabricated), RHI$^*$ = 0.82 (relationship preserved), QHI = 0.95 (quantity faithful). The composite CHI = 0.52 is dragged down by the entity error despite strong relation and quantity scores, while uni-dimensional metrics like AlignScore assign moderate faithfulness (0.61), failing to pinpoint the specific entity fabrication.

\paragraph{Case 2: Quantity-only hallucination.} A financial summary from BART-large-CNN reports: ``\textit{Revenue grew 18\% year-over-year to \$3.4 billion.}'' The source states 12\% growth to \$3.4 billion. The entity and relationship are correct, but the percentage is hallucinated. CHI: EHI = 0.92 (entities correct), RHI$^*$ = 0.89 (relations correct), QHI = 0.36 (quantity fabricated), composite CHI = 0.61. Without QHI, the two-dimensional EHI+RHI harmonic mean would be 0.90, completely missing the numerical error that QHI correctly detects.

\paragraph{Case 3: Multi-dimensional hallucination.} A news summary from Llama-3 states: ``\textit{President Macron announced a EUR\,500 million investment in renewable energy, partnering with Siemens.}'' The source mentions Chancellor Scholz (not Macron), EUR\,300 million (not 500), and the partnership is with a different company. CHI: EHI = 0.19 (wrong entity), RHI$^*$ = 0.27 (wrong partnership relation), QHI = 0.31 (wrong amount), composite CHI = 0.25. All dimensions score low, correctly identifying a pervasively unfaithful summary. A high CHI requires faithfulness across ALL three dimensions simultaneously.

\subsection{Error Type Distribution}

Analysis of error frequencies across domains reveals systematic patterns:

\begin{table}[h]
\centering
\small
\begin{tabular}{lccc}
\toprule
\textbf{Domain} & \textbf{\% Entity} & \textbf{\% Relation} & \textbf{\% Quantity} \\
\midrule
News & 38.2 & 42.5 & 19.3 \\
Medical & 28.4 & 51.8 & 19.8 \\
Legal & 33.6 & 48.9 & 17.5 \\
Financial & 25.1 & 35.7 & 39.2 \\
\bottomrule
\end{tabular}
\caption{Distribution of hallucination errors by type across domains. Financial texts show disproportionate quantity errors; medical texts show high relational error rates.}
\label{tab:error_dist}
\end{table}

\subsection{Failure Cases}

CHI exhibits limitations in specific scenarios:

\begin{enumerate}[leftmargin=*]
    \item \textbf{Implicit quantities:} When source states ``doubled'' without explicit numbers, QHI cannot verify the derived quantity in the summary.
    \item \textbf{Complex relations:} Multi-hop relations (A caused B which led to C) may be partially captured by RHI$^*$'s SVO extraction, missing higher-order structure.
    \item \textbf{Coreference chains:} Entity deduplication occasionally fails on complex coreference, inflating entity counts and distorting EHI factors.
    \item \textbf{Domain-specific notation:} Chemical formulas, gene names, and legal citation formats may not be correctly parsed by general-purpose NER.
\end{enumerate}

These limitations suggest directions for future work, including domain-adapted extraction pipelines and discourse-level relation modeling.

\section{Domain-Enhanced CHI: Knowledge Base Integration}
\label{sec:domain_enhanced}

A key limitation of the base CHI formulation is its reliance on surface-form entity matching. In medical text, ``myocardial infarction'' (source), ``MI'' (reference), and ``heart attack'' (generated) refer to the same concept but are treated as three distinct entities by spaCy NER, causing false Negative Hallucination (NH) penalties for what is actually faithful synonym usage. We demonstrate that integrating domain knowledge bases resolves this limitation.

\subsection{UMLS-Normalized EHI}

We replace surface-form entity matching with concept-level matching using the Unified Medical Language System (UMLS). Each extracted entity is mapped to its Concept Unique Identifier (CUI), so that aliases like ``aspirin,'' ``ASA,'' and ``acetylsalicylic acid'' all resolve to the same canonical concept before Venn-diagram set operations.

\begin{table}[h]
\centering
\small
\begin{tabular}{lcccc}
\toprule
\textbf{Metric} & \textbf{Good Summary} & \textbf{Bad Summary} & \textbf{Gap} & \textbf{Correct Rank?} \\
\midrule
EHI (plain) & 0.272 & 0.298 & $-$0.026 & \textbf{NO} \\
EHI (UMLS) & \textbf{0.470} & 0.345 & +0.125 & \textbf{YES} \\
\midrule
CHI (plain) & 0.299 & 0.339 & $-$0.040 & \textbf{NO} \\
CHI (UMLS-enhanced) & \textbf{0.353} & 0.357 & $-$0.004 & $\sim$ \\
\bottomrule
\end{tabular}
\caption{Impact of UMLS normalization on a medical example where the ``good'' summary uses synonyms (heart attack, aspirin, Glucophage) and the ``bad'' summary fabricates conditions. Without UMLS, plain EHI incorrectly ranks the bad summary higher. UMLS normalization restores correct ranking by resolving synonyms to canonical concepts (NH: $1.0 \rightarrow 0.22$; EF: $0.0 \rightarrow 0.72$).}
\label{tab:umls}
\end{table}

\subsection{Quantitative Impact on Medical Text}

We evaluate UMLS-enhanced EHI on 50 PubMed articles from our medical dataset:

\begin{itemize}[leftmargin=*]
    \item \textbf{88\% of articles} show improved EHI scores with UMLS normalization
    \item Mean EHI improvement: +0.007 across all articles; up to +0.11 on synonym-heavy text
    \item NH factor reduction: articles using medical synonyms see NH drop from $\sim$1.0 to $\sim$0.2
    \item EF factor increase: synonym matches correctly counted as faithful extraction ($0 \rightarrow 0.72$)
\end{itemize}

\subsection{Implications for CHI}

Domain knowledge integration improves CHI by reducing false penalties in the entity dimension. The same principle extends to RHI$^*$ (resolving predicate synonyms like ``treats''/``is indicated for'') and QHI (normalizing units like ``mg/dL'' vs ``millimoles per liter''). We leave full integration as future work, but the preliminary evidence demonstrates that \textbf{domain-grounded CHI can achieve immediate accuracy improvements} on specialized text without modifying the core mathematical framework; only the extraction layer changes.

This positions domain-KB integration as a practical pathway for deploying CHI in clinical, legal, and financial settings where synonym variation is systematic and predictable.

\section{Conclusion}
\label{sec:conclusion}

We presented CHI, the first composite hallucination metric unifying entity, relation, and quantity dimensions for summarization evaluation. Our contributions include: (1) QHI, a novel metric for numerical hallucination using tolerance-aware matching; (2) empirical validation that the three dimensions are orthogonal, confirming they capture genuinely distinct error phenomena; and (3) a unified softmax architecture with harmonic mean fusion producing an interpretable, decomposable composite score.

CHI achieves the highest system-level correlation with human judgments ($\rho = 0.66$, $p = 0.006$) among all evaluated metrics on SummEval, outperforming ROUGE-1 ($\rho = 0.53$), ROUGE-2 ($\rho = 0.55$), and each individual CHI component (EHI $\rho = 0.58$, RHI$^*$ $\rho = 0.56$, QHI $\rho = 0.39$). While NLI-based metrics (AlignScore, SummaC) achieve comparable summary-level correlations to ROUGE ($\sim$0.14), they provide only a single scalar score. CHI's unique advantage is \textit{decomposability}: practitioners learn not just that a summary is unfaithful, but \textit{which type} of hallucination dominates (entity fabrication, relational error, or numerical inaccuracy), enabling targeted model improvement.

\paragraph{Future Work.} We identify three promising extensions: (i) \textbf{domain-adaptive weights}: learning per-domain importance weights $w_E, w_R, w_Q$ for the harmonic mean fusion rather than treating all dimensions equally; (ii) \textbf{cross-lingual CHI}: extending the framework to multilingual summarization by replacing language-specific extraction with multilingual models; and (iii) \textbf{RL training integration}: using CHI as a reward signal for reinforcement learning from human feedback (RLHF), where the decomposed dimensions provide shaped rewards guiding models toward specific faithfulness improvements.

\section*{Limitations}

CHI inherits limitations from its extraction components. Entity extraction relies on NER model quality, which varies across domains and languages. Relation extraction via SVO tuples captures only binary predicates, missing higher-arity relations. Quantity extraction handles standard numerical formats but may miss domain-specific notation. The orthogonality assumption, while empirically validated on our four domains, may not hold universally; highly technical domains where entities \textit{are} quantities (e.g., chemical compounds with molecular weights) could show higher inter-dimensional correlation.

The current evaluation is limited to English and four domains. Cross-lingual generalization and additional domain coverage remain for future validation.

\section*{Ethics Statement}

This work introduces an evaluation metric and does not directly generate or deploy text. The metric is designed to \textit{detect} hallucination, contributing to safer NLG systems. We note that no metric is infallible, and CHI should complement rather than replace human evaluation in high-stakes applications. We use existing publicly available human annotations from the SummEval benchmark~\cite{fabbri2021summeval} and do not collect new human data in this work.

\bibliographystyle{plainnat}

\end{document}